\pdfoutput=1

\documentclass[11pt,table]{article}

\usepackage{acl}
\usepackage{amsmath}
\usepackage{times}
\usepackage{latexsym}
\usepackage{graphicx}

\definecolor{lightgray}{gray}{0.9}

\usepackage[T1]{fontenc}

\usepackage[utf8]{inputenc}

\usepackage{microtype}

\title{Do self-supervised speech models develop human-like perception biases?}

\author{Juliette Millet \\
  CoML, ENS/CNRS/EHESS/INRIA/PSL,\hspace{0.5cm} \\
  LLF, University of Paris, CNRS, \\
  CRI, FAN, IIFR, University of Paris,\\
   Paris, France \\
  \texttt{juliette.millet@cri-paris.org} \\\And
  Ewan Dunbar \\
  CoML, ENS/CNRS/EHESS/INRIA/PSL, \\
  Paris, France \\
  University of Toronto, Toronto, Canada \\
  \texttt{ewan.dunbar@utoronto.ca } \\}

\begin{document}
\maketitle
\begin{abstract}
Self-supervised models for speech processing form representational spaces without using any external labels. Increasingly, they appear to be a feasible way of at least partially eliminating costly manual annotations, a problem of particular concern for low-resource languages. But what kind of representational spaces do these models construct?
Human perception specializes to the sounds of listeners' native languages. Does the same thing happen in self-supervised models? We examine the representational spaces of three kinds of state-of-the-art self-supervised models: wav2vec 2.0, HuBERT and contrastive predictive coding (CPC), and compare them with the perceptual spaces of French-speaking and English-speaking human listeners, both globally and taking account of the behavioural differences between the two language groups. We show that the CPC model shows a small native language effect, but that wav2vec 2.0 and HuBERT seem to develop a universal speech perception space which is not language specific. A comparison  against the predictions of supervised phone recognisers suggests that all three self-supervised models capture relatively fine-grained perceptual phenomena, while supervised models are better at capturing coarser, phone-level, effects of listeners' native language, on perception.
\end{abstract}

\section{Introduction}
Recent advances in speech recognition and representation learning show that self-supervised pre-training is an excellent way of improving performance while reducing the amount of labelled data needed for training. For example, for the LibriSpeech dataset \cite{panayotov2015librispeech}, the current best word error rates \cite{xu2021self,zhang2020pushing} are obtained by systems based on the self-supervised wav2vec 2.0 model \cite{baevski2020wav2vec}. Systems using self-supervised pre-training, both using wav2vec 2.0 and using HuBERT \cite{hsu2021hubert,hsu2021hubert2}, show excellent word error rates after having been fine-tuned on only ten minutes of labelled data. 

What is the effect of this self-supervised pre-training? What type of representational spaces are learned by these models?   \citet{lakhotia2021generative} compared wav2vec 2.0, HuBERT, and contrastive predictive coding (CPC: \citealt{oord2017neural,riviere2021towards}) using an ABX discriminability metric \cite{schatz2016abx}, demonstrating that all three models preserve and enhance linguistically relevant speech sound contrasts in the language they are trained on. We build on this work, asking how these representational spaces compare to the perceptual spaces of human listeners, as inferred from behaviour on phone discrimination experiments.

Human listeners develop speech perception biases under the influence of their native languages. For example, Japanese native speakers tend to confuse the English sounds /r/ and /l/ \cite{yamada1990perception} (\emph{\textbf{r}ight} and \emph{\textbf{l}ight} in English will be perceived as the same or very similar), and English native speakers struggle with the French contrast /y/-/u/ \cite{levy2009assimilation}, having difficulty perceiving the difference between words such as \emph{r\textbf{u}e} (/y/: ``street'') and \emph{r\textbf{ou}e} (/u/: ``wheel''). These misperceptions start to show early on in the native language acquisition process: infants older than 6 months exhibit a facilitating effect at discriminating sounds from their native language, but a decline at doing so for some non-native sounds \cite{kuhl2006infants}. As the importance of this improvement for native sounds and this decline for non-native sounds seems to have a positive impact on infants' future language ability \cite{tsao2004speech,kuhl2005early}, having a perceptual space with native language biases is probably essential to perceive and understand correctly native speech in all situations (with environmental noises, speaker change, etc). If our goal is to have speech models that are as resilient and as adaptable as humans, it is thus interesting to see if they present the same native language specific biases.

By measuring human listeners' ability to discriminate a variety of familiar and unfamiliar speech sounds, we can create a detailed profile of listeners' perceptual biases in the form of a set of sounds' discriminabilities. We then ask whether the training language influences self-supervised speech models in the same way that human listeners' native languages do.

In order to study speech models' perception biases and compare them with humans', we use the Perceptimatic benchmark datasets,\footnote{\url{https://docs.cognitive-ml.fr/perceptimatic/}} a collection of experimental speech perception data intended to facilitate comparison with machine representations of speech. As of this writing, Perceptimatic contains French- and English-speaking participants' behaviour on discrimination tasks for phones in six different languages, for a total of 662 phone contrasts, along with the sound stimuli used during the experiments. 

As in \citet{lakhotia2021generative}, we test state-of-the-art self-supervised models: wav2vec 2.0 \cite{baevski2020wav2vec}, HuBERT \cite{hsu2021hubert,hsu2021hubert2} and a CPC model \cite{riviere2021towards}. We train these models on English and French speech recordings (the native languages of the participants in Perceptimatic). We compare the performance of these self-supervised models with a supervised ASR model, DeepSpeech \cite{amodei2016deep}, trained on the same data but using phonemic labels. To study the degree to which the models' representational space is impacted by properties of speech per se, we also train the same models on recordings of acoustic scenes not including human vocalisations (environmental noises, animal sounds, music, and so on). We use mel-frequency cepstrum coefficients (MFCCs) as an acoustic baseline.

We show that: (1) Self-supervised models trained on speech recordings are better than models trained on acoustic scenes (non-speech) to discriminate speech sounds and to predict human discrimination behaviour (2) They are good at predicting human discrimination behaviour at the stimuli level, but they are worse than neutral acoustic features when we average human results per contrast (3) They show very few native (training) language effect.


All our code and data are freely available.\footnote{\url{https://github.com/JAMJU/Sel_supervised_models_perception_biases}}

\section{Related work}
We are not the first to compare speech models' representational spaces with humans. \citet{feather2019metamers} used metamers as a tool to compare deep neural networks with humans. In a comparison between three speech recognition models, including a fine-tuned wav2vec 2.0 model, \citet{weerts2021psychometrics} showed that wav2vec 2.0 was the best at matching human low-level psycho-acoustic behaviour. However, the model exhibited clear differences with respect to humans---showing, for example, heightened sensitivity to band-pass filtering and an under-reliance on temporal fine structure. 

To perform a comparison at a slightly higher level of speech perception, \citet{scharenborg2018visualizing} visualised a supervised ASR model's internal representations of different speech sounds to investigate its adaptation to new ambiguous phone categories and compare it to humans' behaviour. 

Multiple datasets containing human behavioural data have been collected and openly released to encourage comparison of models with humans. It is for this reason that the Interspeech 2008 Consonant Challenge \cite{cooke2008interspeech} and the OLLO database \cite{meyer2010human}, containing humans' phone identification behaviour in different paradigms, were created. This is also the case for the datasets making up the Perceptimatic database \cite{millet2019comparing,Millet2020PerceptimaticAH,Millet2020ThePE,milletconll} that we employ in this article, which were individually used to study less well-performing models than the ones we use here.

More than just informing us on the kind of information speech models learn, comparing them with humans can have a broader impact on our knowledge of how human perceive speech, and how they learn to do so. \citet{schatz2021early} showed, for example, that a simple self-supervised speech model reproduces the reduced sensitivity to the English [r]/[l] contrast when trained on Japanese speech recordings. Pointing to the fact that the model used lacks abstract phone categories, the authors proposed an alternative to standard explanations of early phonetic learning in infants, as theories about this phenomenon rely heavily on the notion of phone categories. 

With a similar method, \citet{matusevych2020evaluating} tested the ability of various self-supervised speech models to reproduce infants' discrimination behaviour in multiple languages for a small set of pairs of sounds. However, no quantitative comparison with behavioural data was made. Within the same test framework, \citet{schatz2018neural} showed that a neural network trained to perform phone recognition was better at qualitatively reproducing Japanese and English native speakers' discrimination behaviour than an HMM-GMM model, focusing once again on the [r]/[l] pair of sound and also on vowel length differences. In this paper, we decide to: (i) evaluate different self-supervised speech models on more contrasts than these previous works (ii) directly compare their results with human behaviour (iii) measure models' similarity to humans at the \emph{stimuli} level on top of doing it at the contrast level.

\section{Methods}
\subsection{Human ABX test}
Our probes of human speech perception use ABX phone discrimination tests, in which participants hear three speech extracts: A, B and X (an A/B/X \textbf{triplet}). A and B always differ in exactly one phone, and X is always (a distinct recording of) the same sequence of phones as either A or B (for example, A: /pap/, B: /pip/, X: /pap/). We ask the participants to indicate 
which of the first two sounds (A or B) is the most similar to the last sound (X). The ability of the participants to select the correct (\emph{target}) rather than the distractor (\emph{other}) speech extract indicates how well the population tested can discriminate the two phone categories $p_1$ and $p_2$ that \emph{target} and \emph{other} belong to (in our example, /i/ and /a/). We call $p_1$:$p_2$ a \textbf{contrast}. In this paper, we examine the results of monolingual French- and English-speaking participants.

\subsection{Using models to predict}
\label{sec:delta-ABX-exp}
As in previous works \cite{millet2019comparing,Millet2020PerceptimaticAH,Millet2020ThePE,milletconll},  to test models in the same way as participants, we extract a representation $M$ for each of the three stimuli making up each A/B/X triplet in the experiment. We compute, for a triplet \textbf{target}/\textbf{other}/X, each model's $\Delta$-value:
\begin{align}
    \Delta = DTW(M_{other}, M_X) - DTW(M_{target}, M_X)
\end{align}
with $DTW$ being a distance obtained using dynamic time warping to aggregate a frame-level cosine distance along the warping path. The larger (more positive) the $\Delta$-value obtained, the better the model is at discriminating the \textbf{target} and \textbf{other} phone categories. In our comparison between humans' and models' discrimination behaviour, we will generally use the raw $\Delta$-values. The accuracy of the model on a specific triplet, independent of human listeners' behaviour, can also be computed by considering the model to be correct if the corresponding $\Delta$ value is greater than zero and incorrect otherwise. Below, we will refer to this objective accuracy as an \textbf{ABX score.}

\subsection{Models}
\label{sec:models}
We compare self-supervised speech models to see if the representational spaces they develop during training on a language resemble humans' perceptual spaces. We choose to test three state-of-the-art self-supervised  models: contrastive predictive coding (CPC), the basis for the current best-performing systems on the Zero Resource Speech Challenge evaluation \cite{dunbar2021zero}; wav2vec 2.0; and a HuBERT model. These last two models obtain excellent word error rates on the task of semi-supervised speech recognition (self-supervised pretraining plus supervised fine-tuning on a small corpus). 

As we use behavioural data from French and English-speaking participants, models are trained on either French or English recordings. To test for the impact of training on speech recordings compared to other types of sounds, we also train the models on recordings of acoustic scenes (non-speech). We choose one specific output layer for each model, using the one that obtains the best result in terms of human similarity.

We use classic acoustic features as a baseline, using the first 13 mel-frequency cepstrum coefficients (MFCCs), calculated using \textsc{librosa},\footnote{https://librosa.org/} with a window of 25\,ms and a stride of 10\,ms. We also train DeepSpeech \cite{amodei2016deep} as a supervised reference. 

\subsubsection{Contrastive predictive coding}
We use a light version of a model that uses contrastive predicting coding (CPC: \citealt{rivire2020unsupervised}). This model is smaller than HuBERT or wav2vec 2.0, as it is only made up of 5 convolutions (the encoder) and one LSTM layer (the sequence model). It is trained using a contrastive loss. For a sequential input $x=(x_1,...x_t,...,x_T)$, at time $t$, given the output of the sequential model, the loss pushes the model to distinguish the $K$ next outputs of the encoder in the future from randomly sampled outputs from another part of $x$. The detailed loss can be found in Appendix \ref{sec:losses}. We use the output of the sequence model as representations for the CPC model.

\subsubsection{Wav2vec 2.0}
We test wav2vec 2.0 \cite{baevski2020wav2vec}. The model is made up of three elements: an encoder, a quantizer, and a decoder. The encoder is made up of five convolutional layers, the quantizer is a dictionary of possible representations, and the decoder is made up of 12 transformer layers. When an input $z$ is given to the quantizer, it outputs the representation $q$ from the dictionary that is the closest to the input. For an input $x$, wav2vec 2.0 uses the encoder to transform it into $z$, which is then quantized into $q$, and in parallel $z$ is directly passed to the decoder to obtain a context representation $c$.

Like the CPC model, wav2vec 2.0 is trained using a contrastive loss $L_m$. Unlike the CPC model, it uses masking. Given a decoder representation of the context around some masked time step $t$, the loss pushes the model to identify the true quantized speech representation $\mathrm{q}_{t}$ from among a set of $K+1$ quantized candidate representations $\tilde{\mathbf{q}} \in \mathbf{Q}_{t}$  including $\mathrm{q}_{t}$ and $K$ distractors uniformly sampled from other masked time steps in the same utterance (see Appendix \ref{sec:losses} for details). We analyse the fifth layer of the decoder.

\subsubsection{HuBERT}
We also test a HuBERT model \cite{hsu2021hubert,hsu2021hubert2}. This model uses exactly the same architecture as wav2vec 2.0 (except for the quantizer, which is not used), but with a different objective. Its training relies on an unsupervised teacher $h$ (in our case, a K-means algorithm) that assigns a cluster label to each frame. Formally, we have $h(X) = Z=[z_1,...z_T]$, with $z_t$ a $C$-class categorical variable. HuBERT is trained to guess this cluster assignment for masked and unmasked frames at the same time. The detailed loss can be found in Appendix \ref{sec:losses}.

The unsupervised teacher $h$ is initially a K-means clustering on MFCCs. After a round of training using this initial teacher, $h$ is replaced by a  K-means model trained on the output of the sixth transformer layer of the model, and training restarts from scratch. We analyse the output of the sixth transformer layer.

\subsubsection{Supervised reference: DeepSpeech}
As a supervised reference system, we test a trained DeepSpeech model \cite{amodei2016deep}. This model is not too intensive to train, is known to obtain reasonable ASR results, and has previously been compared  to human speech perception \cite{Millet2020ThePE,weerts2021psychometrics}. We train it to generate phonemic transcriptions. 

DeepSpeech is composed of two convolutional layers followed by five RNN layers and a fully connected layer. The model is trained using spectrograms as input and a CTC loss, without a language model.
We use representations extracted from the fourth RNN layer of the model, as it seems to give the best results, both in terms of absolute phone discriminability and for predicting human behaviour.

\subsection{Comparing humans and models' perceptual space}
\label{sec:metrics_compa}
In order to compare humans' and models' perceptual spaces, we use two metrics: the \textbf{log-likelihood} ($\ell\ell$) of a binary regression model on the experimental responses, and the \textbf{Spearman’s $\rho$ correlation} between the average of the model's $\Delta$-values and participants' accuracies averaged within each phone contrast. These allow for predictions at two levels of granularity: the discriminability of individual experimental items ($\ell\ell$) and the overall discriminability of pairs of phones ($\rho$). In the default (\textbf{native}) setting, French-trained models are used to predict French-speaking participants' discrimination results, and similarly for English. See below for details.

For each model tested (see Section \ref{sec:models}), we fit a probit regression to predict the binary responses of the participants (coded as correct or incorrect) using as a predictor the $\Delta$ values obtained from the model's representational space. In addition to a global intercept, the regression has other predictors to account for various nuisance factors: whether the right answer was A (1) or B (0); the order of the trial in the experimental list; a categorical predictor for the participant; and another for the Perceptimatic subset the result belongs to. We fit the model with an L1 regularisation (lasso). The $\ell\ell$ is obtained from the fitted regression model: the larger (less negative) the $\ell\ell$, the better the given model's $\Delta$ values predict the experimental data; thus, the more similar the model's representational space is to the perceptual space of the experimental participants.

We complement the log-likelihood metric with a correlation statistic. We compute the Spearman correlation ($\rho$), a correlation between the ranks of participants' accuracies (using their gradient results if available) and models' $\Delta$-values, both averaged at the level of the phone contrast (zero indicates no correlation, one indicates a perfect monotonic relation). This measure averages out effects of individual A/B/X stimuli below the level of the phone contrast.

\subsection{Comparing native language biases}
\label{sec:native-lang-exp}

Beyond global measures of how well models' representational spaces correspond to human listeners' perceptual spaces, we seek to assess how well the models reproduce group differences caused by the participants' native languages. One could think that humans are very good at discriminating all the sounds from their native language, and that they struggle to differentiate all the sounds from other languages. But reality is more complex than that: some contrasts are equally difficult or easy (even if they are not native) to discriminate for different language groups. The only way to study accurately native language biases is to focus on the relative discrimination difficulties shown by different language groups when listening to the same contrasts.



We present a method which evaluates the ability of the models to directly predict the relative difficulty of contrasts across the two language groups we have in the dataset we use. In other words, we measure if the models, when trained on French and English, show the same discrimination behaviour \emph{differences} than French- and English-speaking participants.

We first normalise the $\Delta$ values obtained by each model by dividing by their standard deviation (within model/training condition, across all A/B/X triplets), in order to put the $\Delta$ values on the same scale for the two models. We average the normalised $\Delta$ values by contrast. We then calculate the overall accuracies for each phone contrast in the listening experiment.

We calculate difference scores: for each phone contrast, we subtract an English model's average $\Delta$ values from the average $\Delta$ value for the corresponding French-trained model. We do the same with the English-speaking and the French-speaking participants' contrast-level accuracy scores. This yields a measure of the \emph{native language effect} for each phone contrast, for each model, and similarly for the human participants.

For each model, we compute a Pearson correlation between its contrast-level native language effects and those of human listeners. The closer  the correlation is to one, the better the phone-level native language effects are captured by a given model.

Because this score calculates a native language effect independently for the models and for the participants, it is not susceptible to the same confounds as an approach which would derive the native language effect from a comparison of  two different (and thus not necessarily comparable) models' fit to the data. Note, however, that the approach we propose is restricted to predicting contrast-level effects of native language.

\section{Experiments}
\subsection{The Perceptimatic dataset}
For the human data, we use five experiments from the Perceptimatic benchmark dataset,\footnote{See \url{https://docs.cognitive-ml.fr/perceptimatic/} for access to, and more detailed descriptions of, the data.} containing the results of French- and English-speaking participants results on ABX phone discrimination experiments. Stimuli come from French, English, Brazilian Portuguese, Turkish, Estonian, and German, and test a variety of  contrasts between vowel and consonant sounds, some of which are familiar, and some of which are unfamiliar, to the listeners. The five datasets use different kinds of stimulus triplets, including short three-phone extracts cut from running speech (\textbf{Zero Resource Speech Challenge 2017} and \textbf{Pilot July 2018} datasets), as well as read-speech nonwords, which highlight English consonants and vowels (\textbf{Pilot August 2018}), compare English with French vowels in a crosslinguistic task (\textbf{Cogsci-2019}), or highlight vowel contrasts in a variety of languages (\textbf{WorldVowels}).
The combined dataset contains 4231 distinct triplets (each of which is sometimes presented to participants in the order target/other/X, sometimes in the order other/target/X), which test 662 phone contrasts, and contains data from 259 French-speaking participants and 280 English-speaking participants (not the same participants for all  stimuli). 

\subsection{Models' training}
The speech models we use are trained on 600-hour subsets of either the English or the French CommonVoice datasets \cite{ardila2019common}. To train DeepSpeech as a phone recognizer, the text transcriptions included in CommonVoice are phonemized using eSpeakNG.\footnote{\url{https://github.com/espeak-ng/espeak-ng}} When English-trained models are used to predict English-speaking participants' results and French-trained for French-speaking participants', we refer to the trained models as \textbf{nat-cpc},
\textbf{nat-w2v}, \textbf{nat-hub}, and \textbf{nat-deep}.

To measure the impact of training on speech versus non-speech audio, the self-supervised models are also trained on a 595-hour subset of the Audioset dataset \cite{gemmeke2017audio} containing no human vocalizations.\footnote{A complete list of the labels kept can be found in our github: \url{https://github.com/JAMJU/Sel_supervised_models_perception_biases}} We refer to these models as   \textbf{aud-cpc},
\textbf{aud-w2v}, and \textbf{aud-hub}.

Each dataset is split randomly into train (80\%), test (10\%) and validation (10\%). All  recordings are resampled at 16000Hz and transformed into mono channel using sox.\footnote{\url{http://sox.sourceforge.net/}}

For the CPC model, we use the Facebook Research implementation\footnote{\url{https://github.com/facebookresearch/CPC_audio}} with all the default parameters. We train the model for 110 epochs and take the models that present the best loss on the validation set.

For wav2vec 2.0, we use the Fairseq Base implementation,\footnote{\url{https://github.com/pytorch/fairseq/tree/master/examples/wav2vec}} using the LibriSpeech configuration. As \cite{baevski2020wav2vec}, we train the models for 400k updates and take the model with the best loss on the validation set.

For HuBERT, we also use the Fairseq Base implementation\footnote{\url{https://github.com/pytorch/fairseq/tree/master/examples/hubert}} and the LibriSpeech configuration. We follow all the training settings of \cite{hsu2021hubert}: our first-pass training takes its unsupervised teacher labels from a K-means algorithm with 50 clusters on the MFCCs for 10\% of the training set, training  for 250k updates. We then extract the representation of the training set from the sixth transformer layer and use these representations to train a new K-means with 100 clusters and re-train the model using these categories as the teacher for 450k updates. We use the model with the best loss on the validation set.

We use a PyTorch implementation of DeepSpeech.\footnote{\url{https://github.com/SeanNaren/deepspeech.pytorch}} We train the models for 150 epochs (to reach an overfitting point), saving a checkpoint of the model for each epoch. We then take the checkpoint that produces the best result in terms of Phone Error Rate (PER) on the validation set. We use specaugment \cite{park2019specaugment} to improve the model performance. The French model obtains 7.8\% PER on the French test set and  the English model obtains 22.75\% PER on the English test set.

\section{Results}
In all graphs, statistical significance of comparisons is evaluated by bootstrapping over participants' results ($N=10000$); redundant statistical comparisons are omitted for clarity (i.e. $C>A$ is omitted when $C>B$ and $B>A$). Confidence intervals shown are 95\% bootstrap intervals. 

\subsection{Overall accuracy}
Before using models' representational spaces to predict human discrimination behaviour, we look at how well models discriminate phones in their training language. We use the sign (positive/negative) of the $\Delta$ values to calculate  the objective accuracy of selecting the target phone  (\textbf{ABX scores}). For interpretability, we calculate scores only on the subsets of Perceptimatic containing monolingual English and French stimuli which were presented to listeners in their native language (\textbf{Zero Resource Speech Challenge 2017}, \textbf{WorldVowels}n and \textbf{Pilot August}). Results are shown in Table \ref{table:delta_positivity}. In general, native self-supervised models obtain scores as good as or  better than the supervised reference and human listeners, with a small preference for the \textbf{nat-w2v} model. They show a clear improvement over the corresponding models trained on acoustic scenes (non-speech). Certain datasets present more difficulties for the self-supervised models relative to \textbf{nat-deep}---notably, the English read-speech nonwords (from the \textbf{WorldVowels} and \textbf{Pilot August} subsets). Further details and comparison of ABX scores between native and non-native settings can be found in Appendix \ref{sec:appendix_abx}.

\begin{table}
\centering
\rowcolors{6}{lightgray}{}
\begin{tabular}{lcc|cc|c}
& \multicolumn{2}{c}{Zero} & \multicolumn{2}{c}{Vowels}  & PilotA\\
& Fr & En & Fr & En & En \\\hline
Humans & {0.84} & 0.80 & {0.80} & {0.84} & 0.74 \\\hline
MFCC & 0.76& 0.77        & 0.73        & 0.76        & 0.88 \\
nat-deep & {0.82} & {0.83}& {0.75}& \textbf{0.87} & {\textbf{0.94}} \\\hline
nat-cpc        & {0.85} & {0.85}& 0.67        & {0.83}& {0.85}\\
~~aud-cpc & 0.76& 0.74        & 0.55        & 0.72        & 0.66    \\
nat-w2v   & \textbf{0.88} & {\textbf{0.88}} & 0.71        & {0.83}& {0.84} \\
~~aud-w2v  & 0.76& 0.73        & 0.53        & 0.71        & 0.78    \\
nat-hub     & {0.87} & {0.87}& {\textbf{0.76}} & 0.83        & 0.82   \\
~~aud-hub   & 0.77& 0.78        & 0.57        & 0.77        & 0.74    \\

\end{tabular}

\caption{ABX scores on three subsets of the Perceptimatic dataset, each containing a French and an English subset;  the larger (closer to one) the better. Scores are averages over the per-triplet accuracies.
Models are native-language models, except those trained on AudioSet. Bold scores are the best in the column.}
\label{table:delta_positivity}
\end{table}

\subsection{Predicting human listeners}
\begin{figure}[h]
   \centering
    \includegraphics[trim ={0cm 0cm 0cm 1cm}, clip,scale = 0.5]{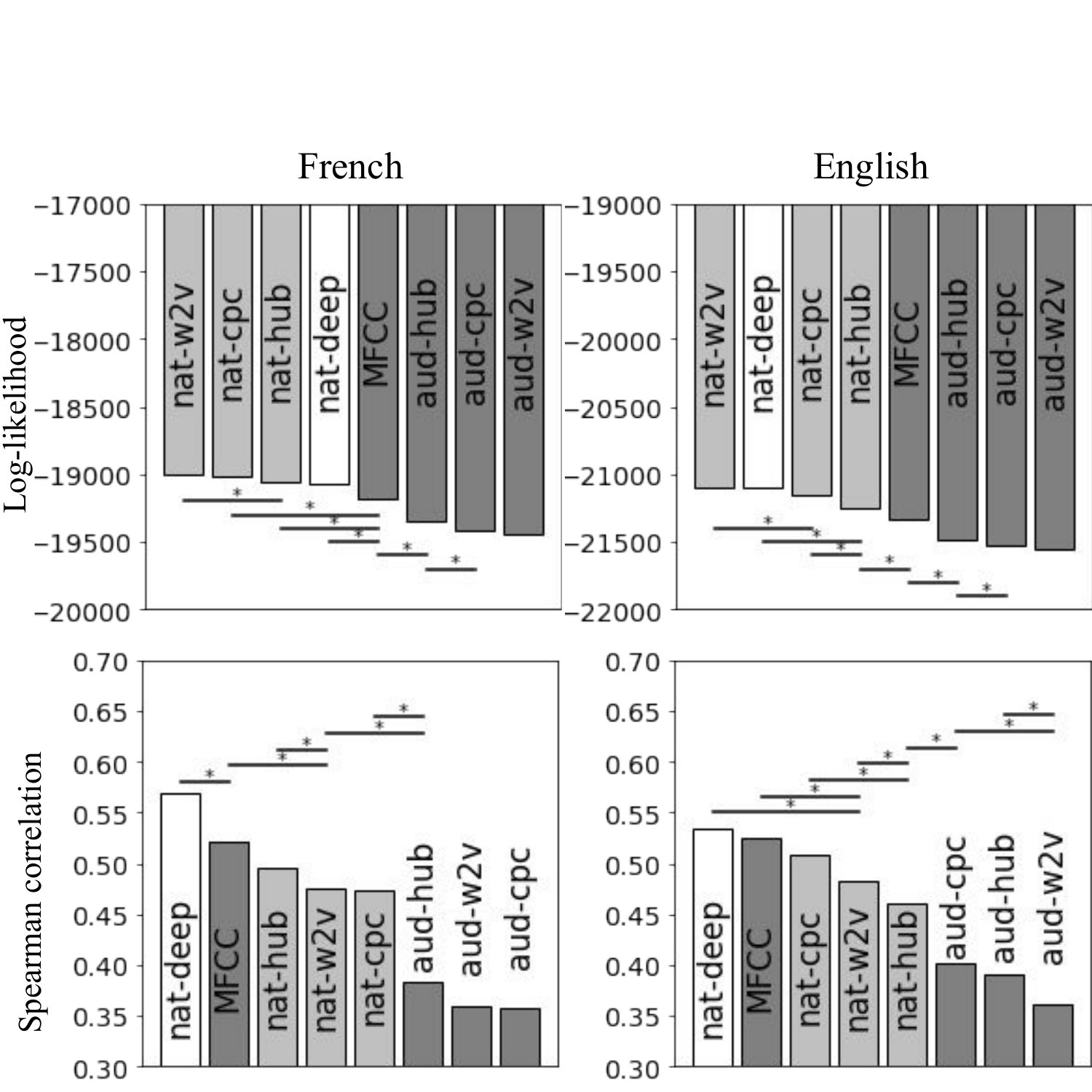}
    \caption{Log-likelihood values (top: shorter/higher bars are better) and Spearman correlation (bottom: taller bars are better) for French (\emph{left}) and English participants (\emph{right}). Stars indicate that the pairwise difference is significant. The supervised reference is in white to distinguish it from the native self-supervised models in light grey and the baselines in darker grey (neutral acoustic features and models trained on acoustic scenes).}
    \label{fig:results_predictions}
\end{figure}

To assess how well self-supervised models' representational spaces match humans' perceptual spaces for speech, we compute the log-likelihood ($\ell\ell$) and the Spearman correlation ($\rho$) metrics over the entire Perceptimatic dataset (see Section \ref{sec:metrics_compa}) in the native-language training condition. Results can be seen in Figure \ref{fig:results_predictions}.

First, we need to note that the models' performance appears to be importantly tied to training on speech, rather than simply on natural audio. Indeed, the models trained on acoustic scenes (non-speech) consistently perform worse than the native-trained models and MFCCs, on both measures.

For the $\ell\ell$ metric, \textbf{nat-w2v} does at least as well as, or (for French) somewhat better than, the supervised reference at modelling human listeners' perceptual confusions; most native self-supervised models perform similarly. Self-supervised models  appear to learn representational spaces at least as similar to human native listeners' as our supervised phone recogniser when measured in this way.

The $\rho$ metric, which correlates models' with humans' average dissimilarity ($\Delta$ or accuracy) for each phone contrast, reveals a different pattern. Here, \textbf{nat-deep} performs best. Furthermore, native self-supervised models perform worse than generic MFCC features. This suggests a component of human speech perception that is poorly captured by self-supervised models at the contrast level. (On some subsets---notably the \textbf{WorldVowels} set of familiar and unfamiliar vowel contrasts---self-supervised models \emph{are} better than MFCCs, but are still worse than our supervised reference; see Appendix \ref{sec:appendix1}.)

\begin{figure}[h]
   \centering
    \includegraphics[trim ={0cm 0cm 0cm 0cm}, clip,scale = 0.51]{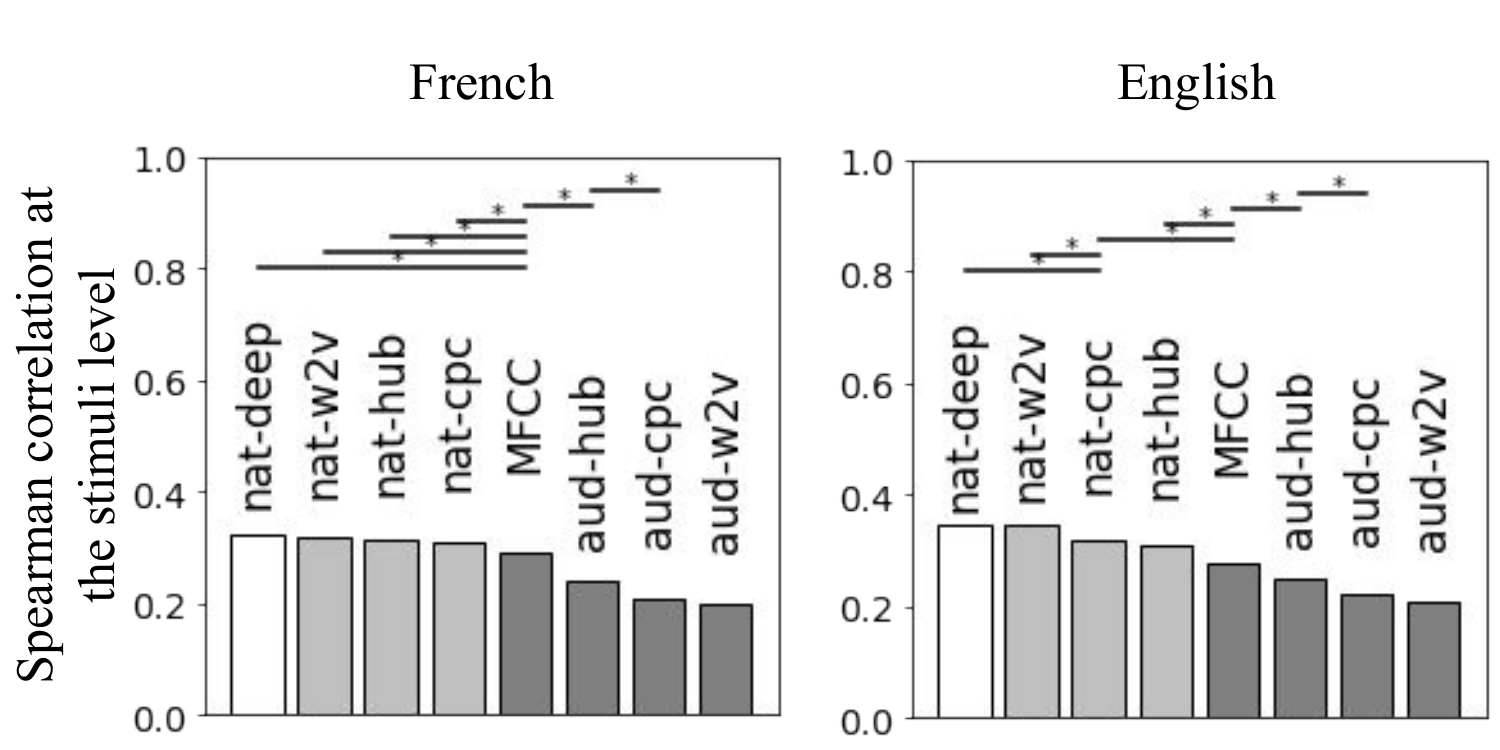}
    \caption{Spearman correlation at the stimuli level (taller bars are better) for French (\emph{left}) and English participants (\emph{right}). Stars indicate that the pairwise difference is significant. }
    \label{fig:spear_comp}
\end{figure}

To confirm the difference of result for the contrast level (the $\rho$ metric) and the stimuli level (the $\ell\ell$ metric), we compute the Spearman correlation metric at the stimuli level, averaging participants' results over the stimuli, instead of doing it, for models and humans, over contrasts. The results of this analysis can be found in Figure \ref{fig:spear_comp}. We notice that this new analysis, done at the stimuli level, gives similar results than our log-likelihood metric. This supports the idea that the bad results for the original $\rho$ metric of the self-supervised models we consider are due to the averaging over contrast.

\begin{figure}[h]
    \centering
    \includegraphics[scale=0.115]{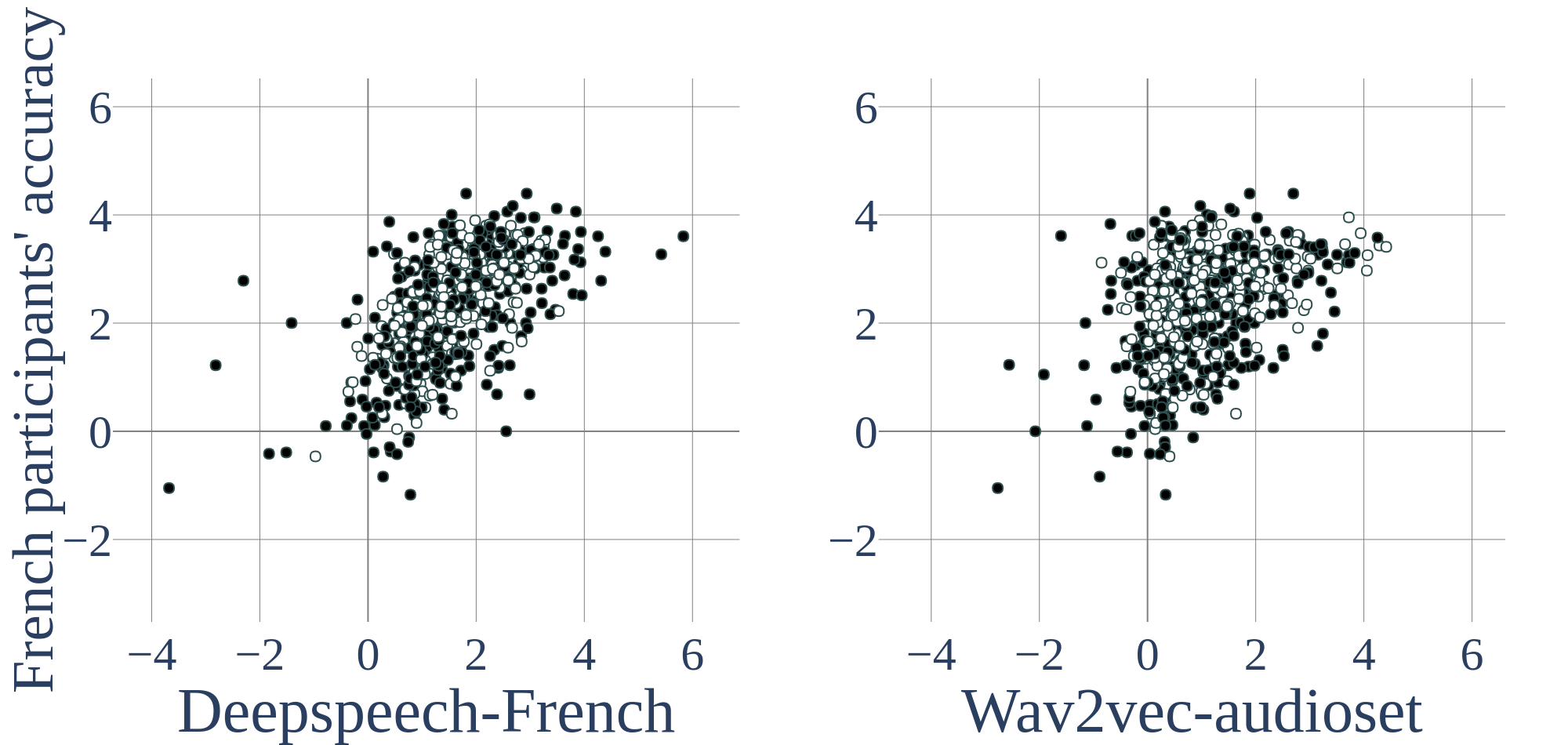}
    \caption{Average of French listeners' results (higher: better discrimination) against average $\delta$ from (\textbf{left}) supervised reference trained on phonemic transcriptions (\textbf{right}) wav2vec trained on non-speech recordings. Each point is a contrast. Measures are normalised by dividing by standard deviation over the entire data set, so the two scales are comparable. Black circles are non-native contrasts, white ones are native (French).}
    \label{fig:deep_audio}
\end{figure}

To illustrate the comparisons at the level of phone contrasts, in Figure \ref{fig:deep_audio} we plot the average accuracy (per contrast) for French-speaking participants results against (left) DeepSpeech trained on French, one of the best-performing models, and (right) wav2vec 2.0 trained on AudioSet (aud-w2v), one of the models that is the least similar to humans.
 
\subsection{Native language biases}

\begin{figure}[h]
   \centering
    \includegraphics[trim ={0cm 0cm 0cm 0cm}, clip,scale = 0.5]{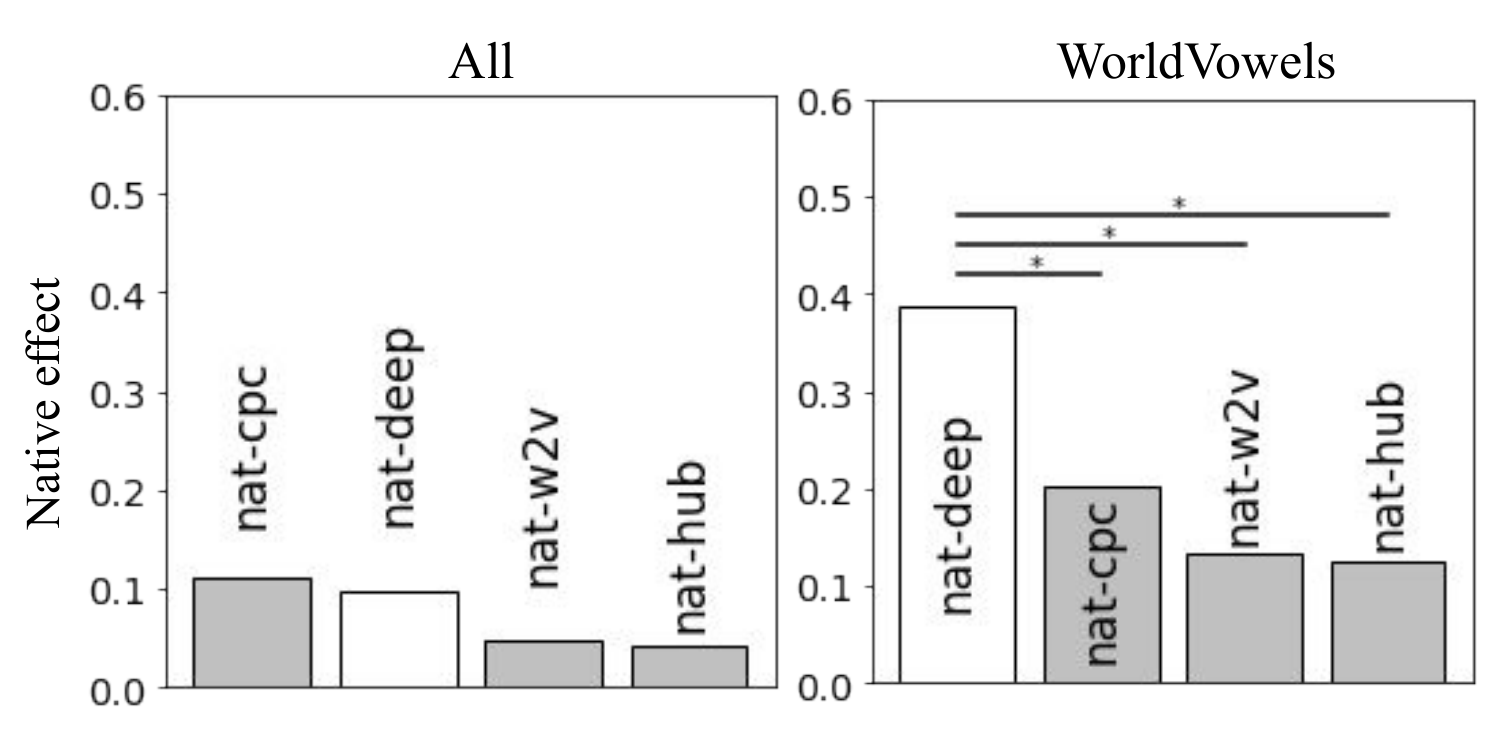}
    \caption{Native language effect for each model, the bigger the bar, the better the models capture language specificities in the discrimination behaviour between the two  groups. Stars indicate that the pairwise difference is significant. The supervised reference is in white to distinguish it from the self-supervised models in light grey.}
    \label{fig:results_native}
\end{figure}

To look for the presence of human-like native language biases, we look at the ability of native models to predict the difference in behaviour between the French- and the English-speaking groups (see Section \ref{sec:native-lang-exp}). Figure \ref{fig:results_native} (left) shows the native language effect assessed over the entire Perceptimatic dataset---that is, the correlation, at the contrast level, between the differences in $\Delta$ across language-training conditions, on the one hand, and the differences in accuracy for the two listener groups, on the other. \textbf{Nat-cpc} is competitive with \textbf{nat-deep} at predicting differences in groups. \textbf{Nat-hub} and \textbf{nat-w2v}, on the other hand, show very native language effect. 

Figure \ref{fig:results_native} (right) shows the same analysis, but on only the \textbf{WorldVowels} dataset. The stimuli in this dataset are constructed to specifically induce different discrimination behaviour between the two language groups. Here, \textbf{nat-deep} shows a much better ability to predict native language effects, both in the absolute, and relative to the other models. 

\begin{figure}[h]
   \centering
    \includegraphics[trim ={0cm 0cm 0cm 0cm}, clip,scale = 0.5]{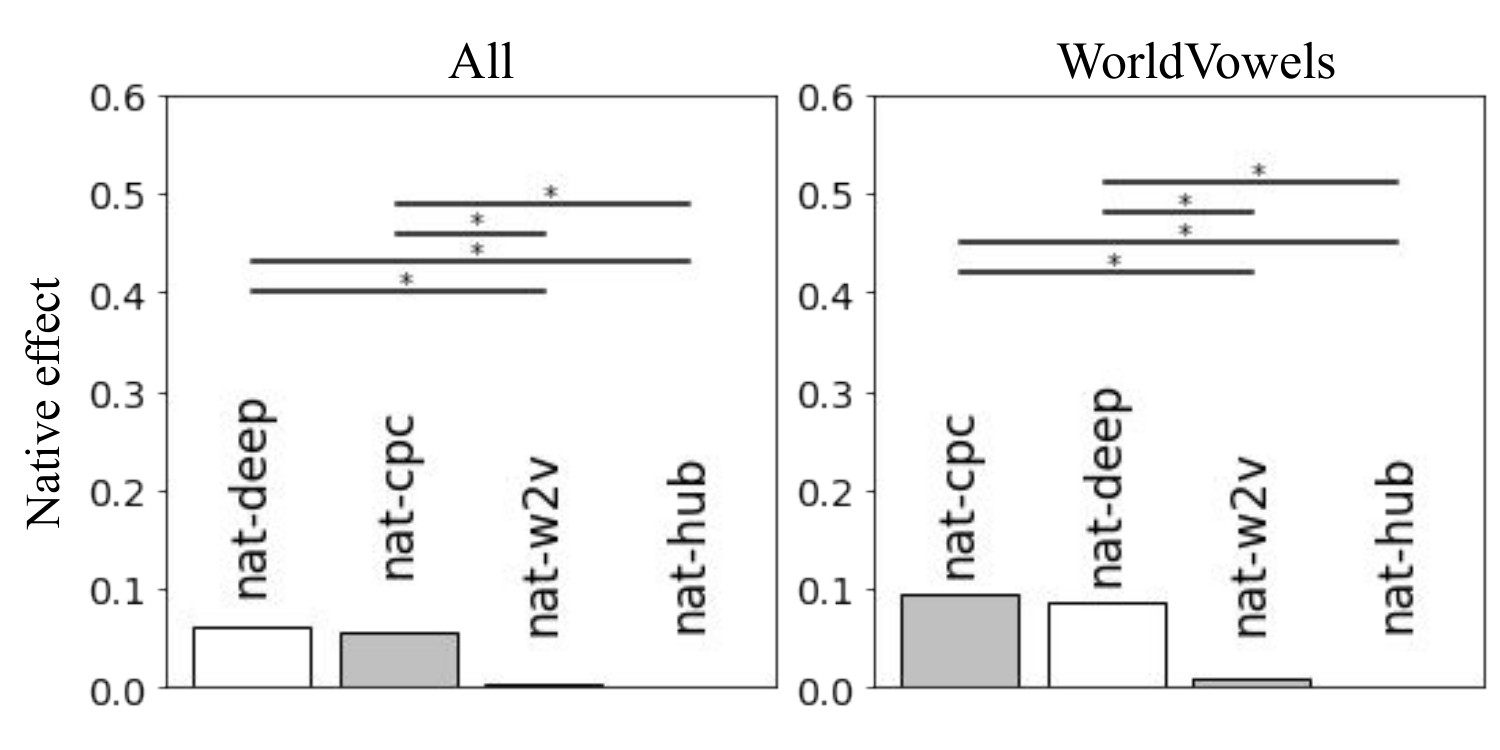}
    \caption{Native language effect for each model, the bigger the bar, the better the models capture language specificities in the discrimination behaviour between the two  groups. Stars indicate that the pairwise difference is significant. The supervised reference is in white to distinguish it from the self-supervised models in light grey.}
    \label{fig:native_file}
\end{figure}

As this analysis is done at the level of phone contrasts, and not individual stimuli, we could think that as our supervised reference model is trained to produce phonemic transcriptions, it probably gives it a head start at predicting differences in discrimination behaviour driven by phone categories. To look more precisely at this, we compute our native effect at the stimuli level instead of the contrast level. The results of this analysis can be seen in Figure \ref{fig:native_file}, for the all dataset and for the WorldVowels subset. Going to the stimuli level reduces radically the native effect measured. This is expected, as the number of participants' result per stimulus is small, and the effect measured on humans is thus very noisy when measured at this level, and therefore harder to reproduce for the models. However, we can notice that our supervised reference and the CPC model are still the ones that exhibit the most native language effect.

\section{Discussion}

We showed that the self-supervised models we tested seem to learn representational spaces relevant for predicting human phone discrimination at the stimuli level. However, while humans show consistent discrimination behaviour for certain contrasts, whatever the stimuli, the self-supervised models we test do not capture systematic effects of contrasts between specific pairs of phones. Unlike our supervised reference, their similarity to human perceptual spaces is limited to capturing the discriminability of specific individual stimuli. The models tested were similar, but wav2vec 2.0 showed a slight advantage for predicting this kind of behaviour.

We have also shown that training on speech data is essential to obtaining a human-like perceptual space: for all of our metrics (ABX accuracy or similarity to humans), training on speech leads to better results than training on acoustic scenes (non-speech). This strongly suggests that the benefits of self-supervised speech models comes from learning characteristics of human speech, not simply the fact that they are better general audio features. We speculate that this is not just important to their ability to predict human speech perception and to discriminate phones, but also of their (related) utility for doing downstream tasks such as ASR.

What these models learn about speech, however, is not typically language-specific---at least, not in the same way that human perception is. Wav2vec 2.0 and HuBERT do not model language-specific differences in human speech perception, and can be seen as modelling a language-neutral or universal speech perception space. Indeed, they exhibit very few native language effect (see Figure \ref{fig:results_native} and \ref{fig:native_file}).  We note that the idea of self-supervised models learning universal speech features is consistent with the fact that models trained on one language, or multilingually, have proven useful for representing speech in unseen languages \cite{riviere2020unsupervised}.

CPC does capture effects of native language on perception at the contrast level, but to a far lesser extent than our supervised reference when we focus on a subset of Perceptimatic designed to capture important differences in discrimination behaviour for our two groups of participants (WorldVowels). Our CPC model differs from the other models tested in its small size, its causal architecture (wav2vec and HuBERT use transformers), and in that it does not use masking during its training. Its architecture is probably the most biologically plausible of the three self-supervised models we tested. We should note, however, that it does not make it the best predictor of human discrimination behaviour among the three models (see Figure \ref{fig:results_predictions} and \ref{fig:spear_comp}). 

One possible explanation for the self-supervised models' limitations we observe is insufficiency of training data: the models in question have generally shown good performance on downstream tasks when pre-trained on large amounts of data. We tested this using available pretrained wav2vec and HuBERT models trained on much larger amounts of data. The detailed results can be found in  Appendix \ref{sec:pretrained}. The models show a slight improvement, but, when looking at the $\rho$ statistic at the phone contrast level, they are still worse than MFCCs.

Contrary to previous results \cite{Millet2020PerceptimaticAH,Millet2020ThePE}, our supervised reference system is quite good at predicting human discrimination behaviour (in particular at the contrast level), and clearly predicts a native language effect. The main differences in our experiment with \cite{Millet2020ThePE} are the type of model (DeepSpeech instead of HMM-GMM), and with \cite{Millet2020PerceptimaticAH} the type of training objective (phone recognition rather than prediction of orthographic text), and the size of the training corpora (we use fewer data). Predicting phones rather than orthography seems to be critical (as we demonstrate in Appendix \ref{sec:DeepSpeech_ortho}), and using a neural network instead of a Bayesian model (HMM-GMM) leads to a more human-like representational space, as already highlighted by \cite{schatz2018neural}. 

Given the advantage supervised phone recognizers show, a different approach to developing more human-like representational spaces in self-supervised models might be the inclusion of tasks or constraints that push them to take into account longer time scales in order to encourage them to construct longer, more phone-like units.

   
\section*{Acknowledgements}
This research was supported by the \'Ecole Doctorale Frontières du Vivant (FdV) -- Programme Bettencourt, by the Connaught Fund and  the Arts and Science Tri-Council Bridging Fund, University of Toronto, and by French Agence Nationale de la Recherche grants ANR-17-CE28-0009 (GEOMPHON), ANR-11-IDFI-023 (IIFR), ANR-11-IDEX-0005 (USPC),  ANR-10-LABX-0083 (EFL),  ANR-17-EURE-0017 Frontcog, ANR-10-IDEX-0001-02 PSL*, ANR-19-P3IA-0001 PRAIRIE 3IA Institute. This work was performed using HPC resources from GENCI-IDRIS (Grant 2021-AD011012415, 2022-AD011012415R1).

\bibliography{anthology,custom}
\bibliographystyle{acl_natbib}
\appendix

\section{Detailed losses used by the models}
\label{sec:losses}
The loss used by the \textbf{CPC} model is the following:

\begin{align}
    \mathcal{L}_{t}=-\frac{1}{K} \sum_{k=1}^{K} \log \left[\frac{\exp \left(\phi\left(\mathbf{x}_{t+k}\right)^{\top} \mathbf{A}_{k} \mathbf{z}_{t}\right)}{\sum_{\mathbf{n} \in \mathcal{N}_{t}} \exp \left(\phi(\mathbf{n})^{\top} \mathbf{A}_{k} \mathbf{z}_{t}\right)}\right]
\end{align}
Where $A_k$ is a learned linear classifier, $\phi$ is the encoder, and $N_i$ is the set of negative examples. With an input $x_t$ and an output $\mathbf{z}_{t}=\psi\left(\phi\left(\mathbf{x}_{1}\right), \ldots, \phi\left(\mathbf{x}_{t}\right)\right)$, with $\psi$ the sequential model, it pushes the model to identify the $K$ next outputs $\phi\left(\mathbf{x}_{t+k}\right)$ in the future, in comparison with randomly sampled outputs from another part of $x$.

The loss used by the \textbf{wav2vec 2.0} model is the following: 
\begin{align}
    \mathcal{L}_{m}=-\log \frac{\exp \left(\operatorname{sim}\left(\mathbf{c}_{t}, \mathbf{q}_{t}\right) / \kappa\right)}{\sum_{\tilde{\mathbf{q}} \sim \mathbf{Q}_{t}} \exp \left(\operatorname{sim}\left(\mathbf{c}_{t}, \tilde{\mathbf{q}}\right) / \kappa\right)}
\end{align}

for a masked time step $t$, the model has to choose the true quantized speech representation $\mathrm{q}_{t}$ in a set of $K+1$ quantized candidate representations $\tilde{\mathbf{q}} \in \mathbf{Q}_{t}$ which includes $\mathbf{q}_{t}$ and $K$ distractors. The model also use a diversity loss so the representation in the quantizer dictionary be as diverse as possible, for more details, see \cite{baevski2020wav2vec}.

The loss used by \textbf{HuBERT} is the following:
\begin{align*}
    L(f;X,M,Z) &=& \alpha \sum_{t \in M} \log p_{f}\left(z_{t} \mid \tilde{X}, t\right) + \\ && (1- \alpha) \sum_{t \notin M} \log p_{f}\left(z_{t} \mid X, t\right)
\end{align*}

With $\alpha \in [0,1]$, M the set of masked frames, f the cluster assignment predictor, and $\hat{X}$ masked frames.
\section{Predicting human results: results on sub-datasets}
\label{sec:appendix1}
\begin{figure}[h]
   \centering
    \includegraphics[trim ={0.2cm 0cm 0cm 1cm}, clip,scale = 0.51]{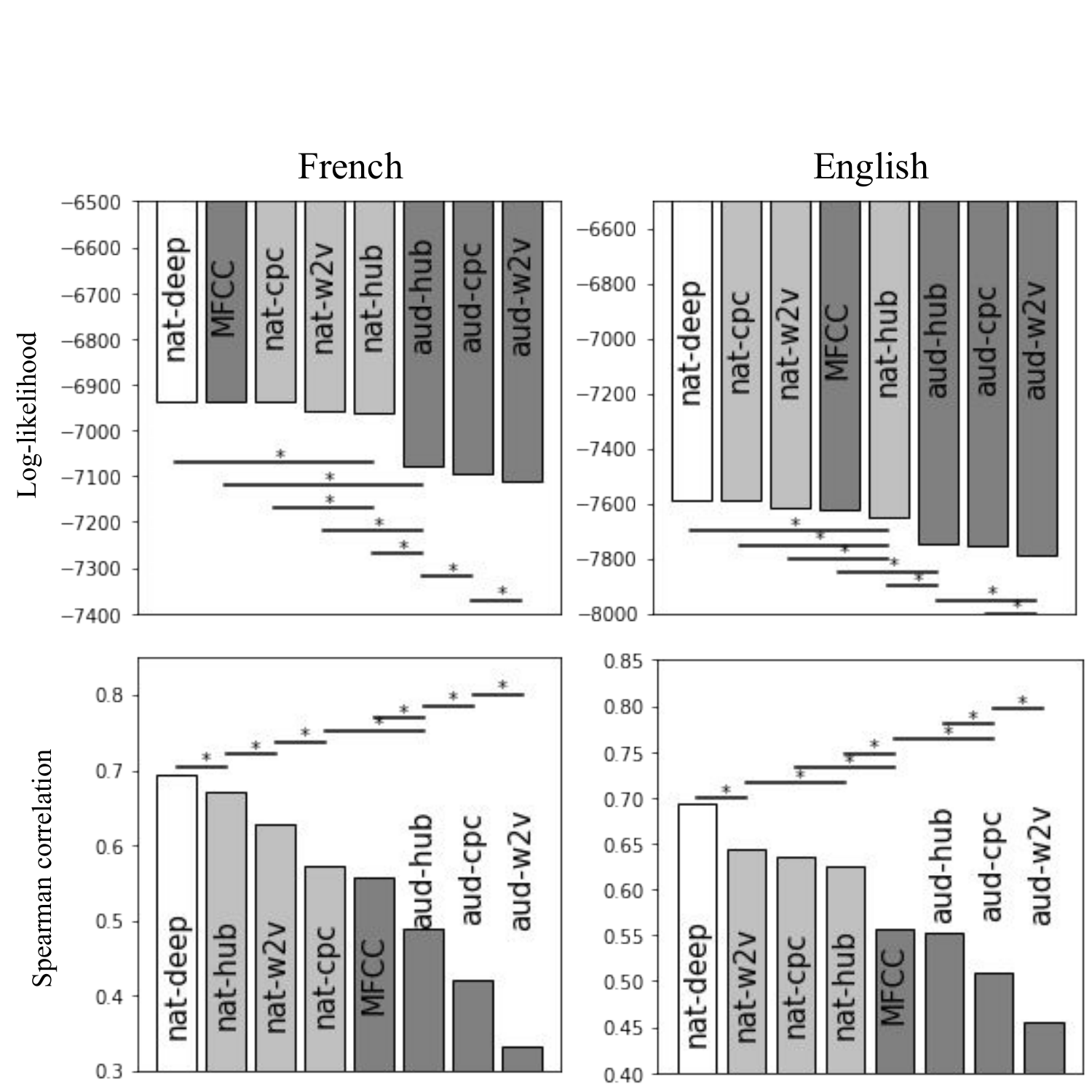}
    \caption{Results on the \textbf{WorldVowels} subset. Log-likelihood values (top: shorter bars are better) and Spearman correlation (bottom: taller bars are better) for French (\emph{left}) and English participants (\emph{right}). Stars indicate that the pairwise difference is significant. The supervised reference is in white to distinguish it from the self-supervised model trained on speech recordings (in light grey), and the baselines in darker grey (neutral acoustic features and models trained on acoustic scenes).}
    \label{fig:results_predworldvowels}
\end{figure}

\begin{figure}
   \centering
    \includegraphics[trim ={0.2cm 0cm 0cm 1cm}, clip,scale = 0.51]{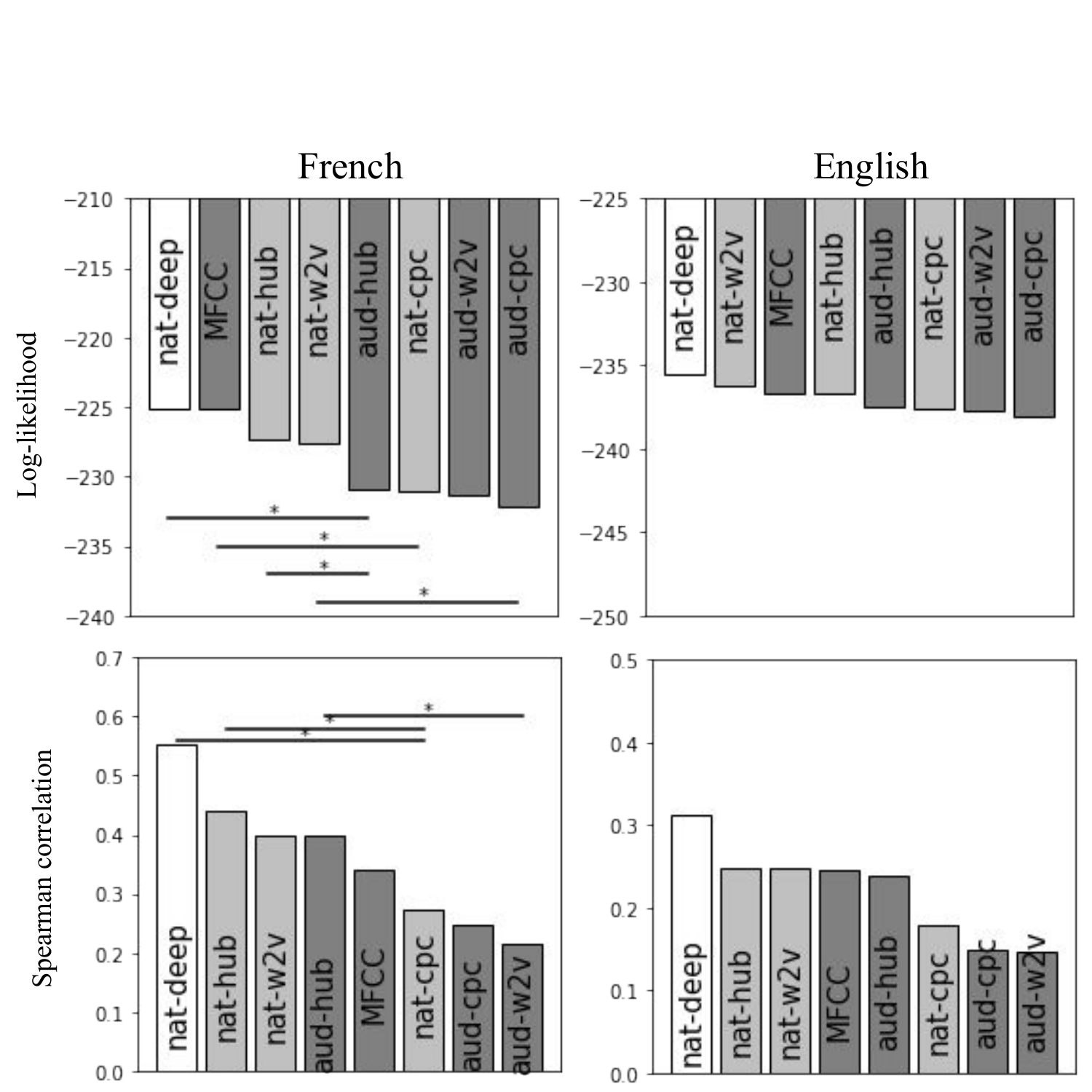}
    \caption{Results on the \textbf{Cogsci-2019} subset. Log-likelihood values (top: shorter bars are better) and Spearman correlation (bottom: taller bars are better) for French (\emph{left}) and English participants (\emph{right}). Stars indicate that the pairwise difference is significant. The supervised reference is in white to distinguish it from the self-supervised model trained on speech recordings (in light grey), and the baselines in darker grey (neutral acoustic features and models trained on acoustic scenes).}
    \label{fig:results_predcogsci}
\end{figure}

\begin{figure}[h]
   \centering
    \includegraphics[trim ={0.2cm 0cm 0cm 1cm}, clip,scale = 0.51]{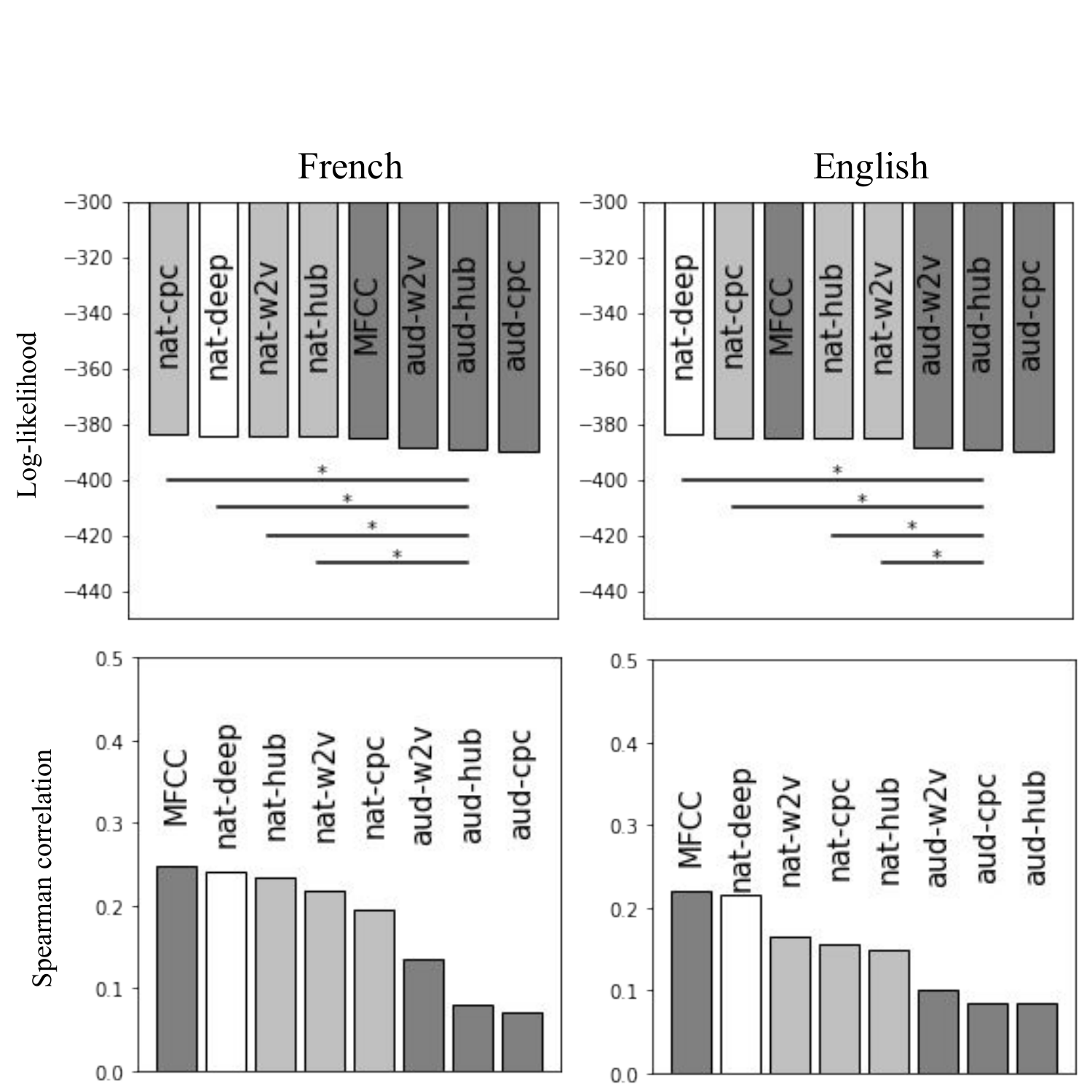}
    \caption{Results on the \textbf{pilot-july-2018} subset. Log-likelihood values (top: shorter bars are better) and Spearman correlation (bottom: taller bars are better) for French (\emph{left}) and English participants (\emph{right}). Stars indicate that the pairwise difference is significant. The supervised reference is in white to distinguish it from the self-supervised model trained on speech recordings (in light grey), and the baselines in darker grey (neutral acoustic features and models trained on acoustic scenes).}
    \label{fig:results_pilotjuly}
\end{figure}

\begin{figure}
   \centering
    \includegraphics[trim ={0.2cm 0cm 0cm 1cm}, clip,scale = 0.51]{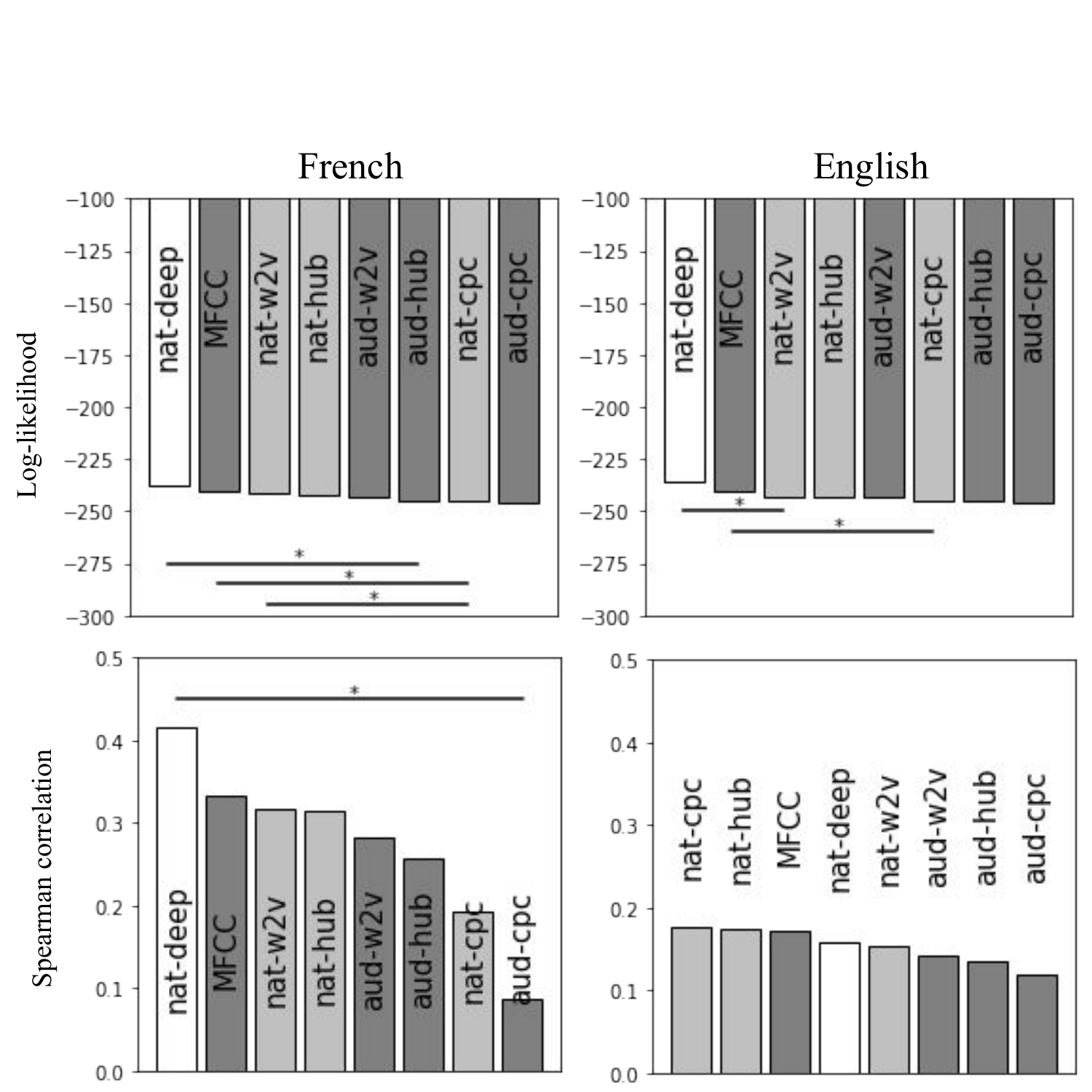}
    \caption{Results on the \textbf{pilot-august-2018} subset. Log-likelihood values (top: shorter bars are better) and Spearman correlation (bottom: taller bars are better) for French (\emph{left}) and English participants (\emph{right}). Stars indicate that the pairwise difference is significant. The supervised reference is in white to distinguish it from the self-supervised model trained on speech recordings (in light grey), and the baselines in darker grey (neutral acoustic features and models trained on acoustic scenes).}
    \label{fig:results_pilotaugust}
\end{figure}

\begin{figure}
   \centering
    \includegraphics[trim ={0.2cm 0cm 0cm 1cm}, clip,scale = 0.51]{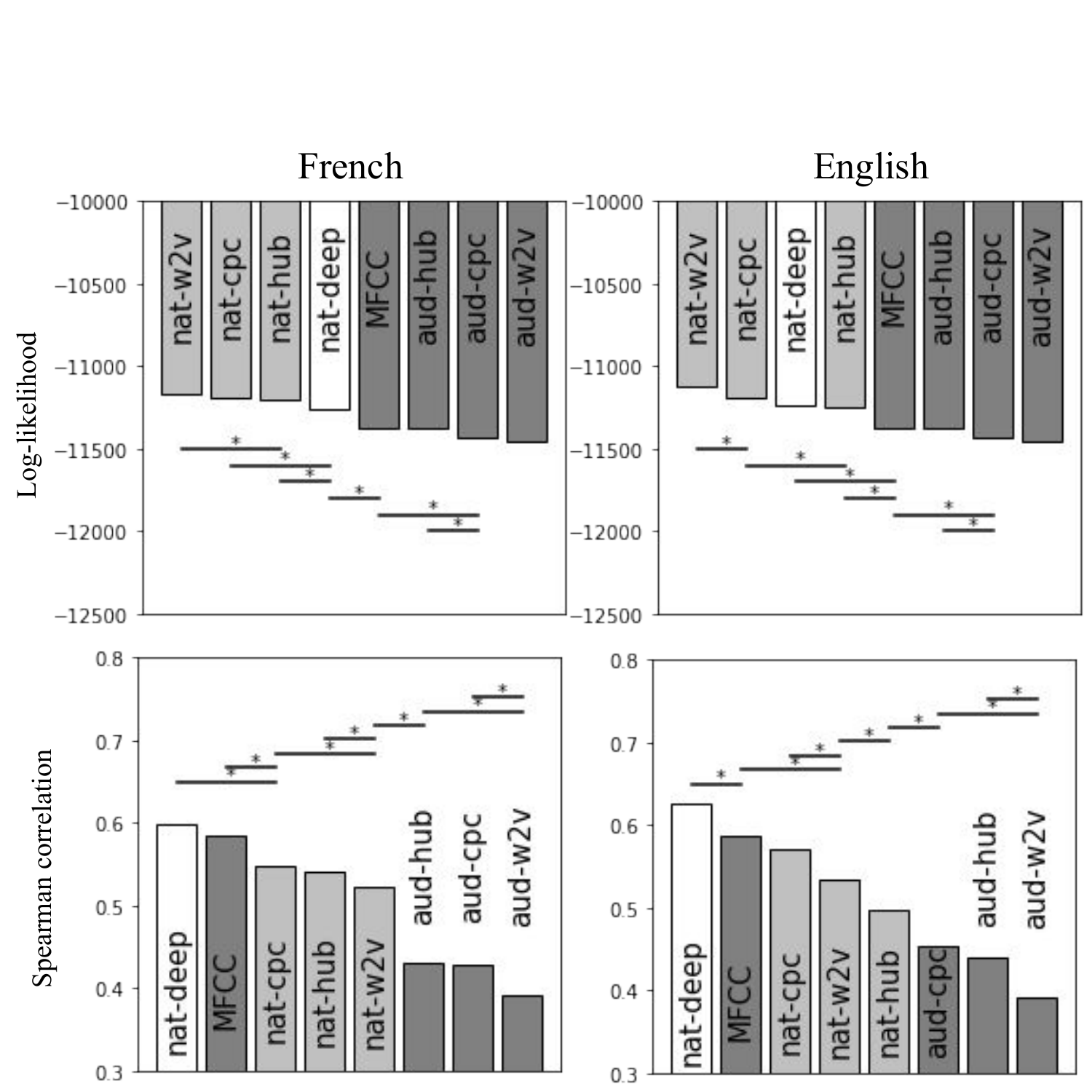}
    \caption{Results on the \textbf{Zerospeech} subset. Log-likelihood values (top: shorter bars are better) and Spearman correlation (bottom: taller bars are better) for French (\emph{left}) and English participants (\emph{right}). Stars indicate that the pairwise difference is significant. The supervised reference is in white to distinguish it from the self-supervised model trained on speech recordings (in light grey), and the baselines in darker grey (neutral acoustic features and models trained on acoustic scenes).}
    \label{fig:results_zerospeech}
\end{figure}

We present the results on the different Perceptimatic subsets. The results for Cogsci 2019 can be seen in Figure \ref{fig:results_predcogsci}, for WorldVowels in Figure \ref{fig:results_predworldvowels}, for Zerospeech in Figure \ref{fig:results_zerospeech}, for pilot-july in Figure \ref{fig:results_pilotjuly}, and for pilot-august in Figure \ref{fig:results_pilotaugust}. These results should be taken carefully, in particular for the Cogsci subset and the pilots, as not much contrasts and stimuli were tested for these subsets compared to the others.

\section{Difference in ABX score between French and English models}
\label{sec:appendix_abx}
\begin{figure}
    \centering
    \includegraphics[scale=0.43]{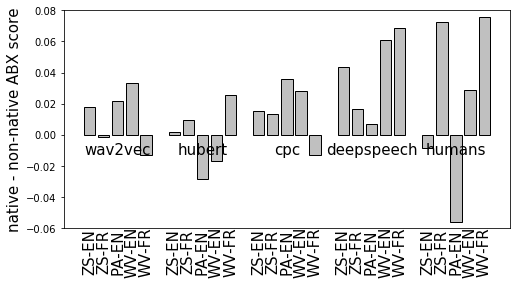}
    \caption{ABX score difference between native setting and non-native setting for the different models tested. The bigger the bar above zero, the bigger difference.}
    \label{fig:ABX_diff}
\end{figure}

To complete Table \ref{table:delta_positivity}, we present in Figure \ref{fig:ABX_diff} the detailed ABX score difference between a native discrimination setting (English models and participants discriminating English contrasts and same for French) and a non-native discrimination setting (English models and participants discriminating French contrasts and vice-versa). Humans' ABX scores differences show that English-speaking participants are not always better than French-speaking participants at discriminating English sounds (for the Zerospeech subsets for example).

\section{Language preference}
\label{sec:appendix_lang_pref}
\begin{figure}
   \centering
    \includegraphics[trim ={0.2cm 0cm 0cm 1cm}, clip,scale = 0.51]{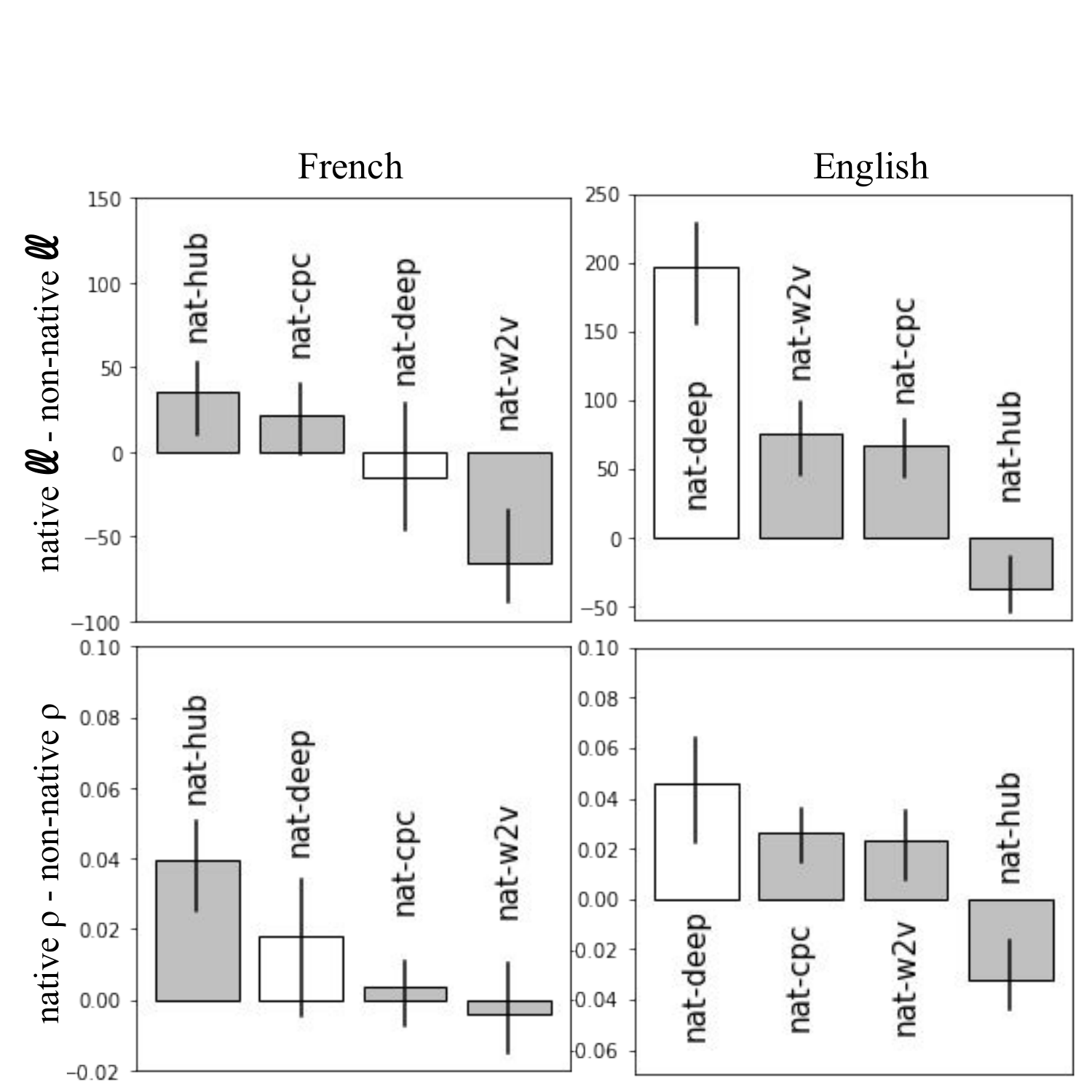}
    \caption{Native minus non-native log-likelihood values (top) and Spearman correlations (bottom) for French (\emph{left}) and English participants (\emph{right}). The higher the bar above zero, the better the native setting is compared to the non-native setting. The supervised reference is in white,  the self-supervised models are in light grey. Black lines indicate 95\% confidence intervals.}
    \label{fig:results_diffnative}
\end{figure}

A possible approach to study models' language specificity would be to see if English-trained models predict English-speaking participants better than French-trained models, and vice versa.
We assess whether models in the native training condition predict discriminability better than the corresponding models in the non-native training condition. Figure \ref{fig:results_diffnative} plots the subtraction of the $\ell\ell$ and $\rho$ scores in the non-native setting from the corresponding scores in the native setting (across the entire Perceptimatic dataset). 

For both the (experimental item-level) $\ell\ell$ and the (phone contrast-level) $\rho$ score, DeepSpeech consistently outperforms over wav2vec 2.0. This is in contrast with the overall prediction performance reported above, where wav2vec 2.0 was on par with DeepSpeech, DeepSpeech generally shows a relative advantage for predicting the behaviour of listeners whose native language is the same as the training language, while wav2vec 2.0 does not.

There is a striking difference between languages in the performance of DeepSpeech: for English, the native DeepSpeech shows a substantial advantage over the non-native (French-trained) DeepSpeech which is not present for the French datasets. Similarly, in French, the native HuBERT shows an advantage over the non-native (English-trained) HuBERT, while the reverse is true in English. However, these two major differences may be in part explained by global effects: the French-trained HuBERT model is better at predicting the results for all participants (not just French-speaking participants), as is the English-trained DeepSpeech model. 


\section{Using pretrained models on more data}
\begin{figure}
   \centering
    \includegraphics[trim ={0.2cm 0cm 0cm 1cm}, clip,scale = 0.51]{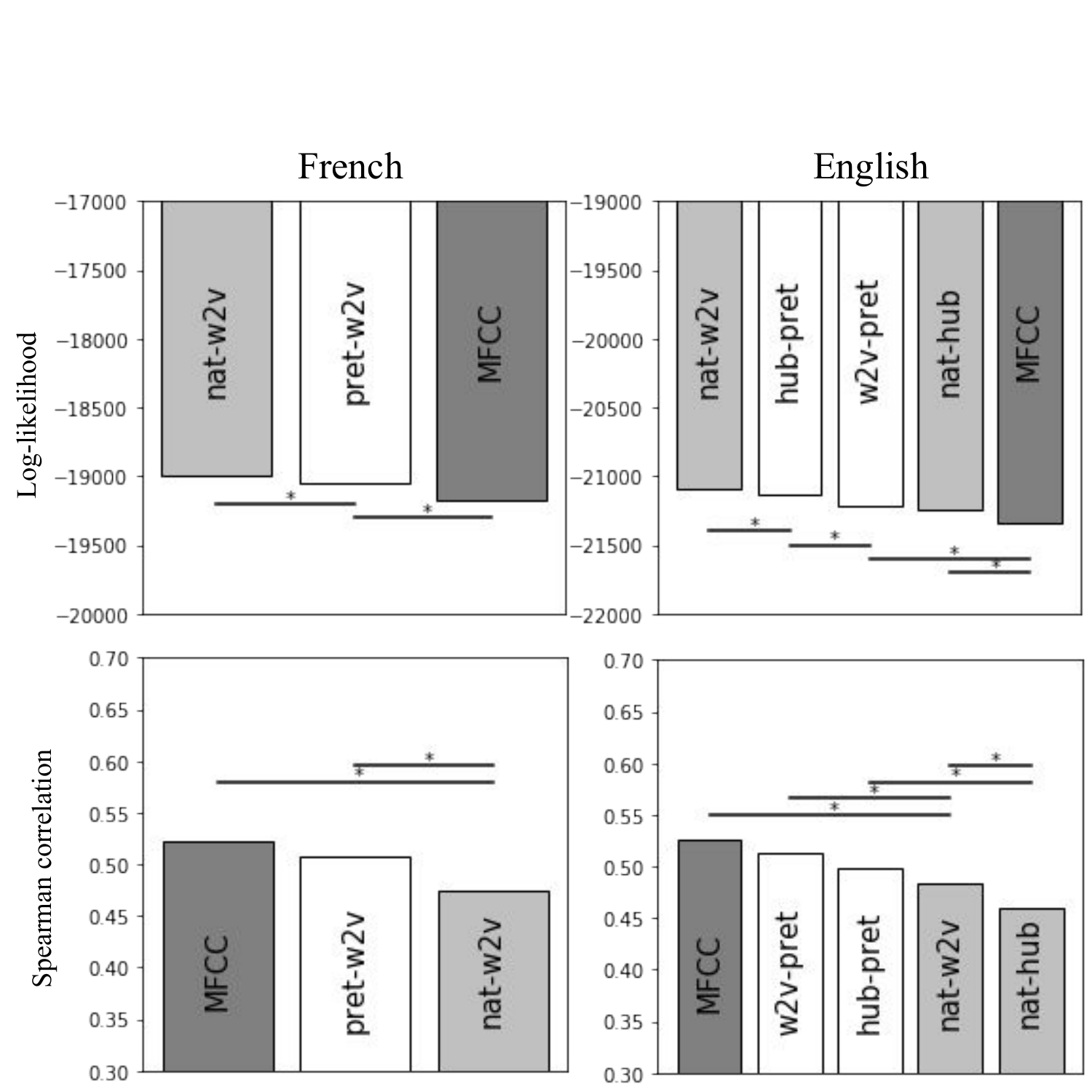}
    \caption{Log-likelihood values (top: shorter bars are better) and Spearman correlation (bottom: taller bars are better) for French (\emph{left}) and English participants (\emph{right}). Stars indicate that the pairwise difference is significant. The pretrained models are in white to distinguish it from our self-supervised models trained on only 600h of speech.}
    \label{fig:pretrained}
\end{figure}

We compare our models with pretrained models available online. 
For English, we tested a wav2vec and a HuBERT model trained on Librispeech \cite{panayotov2015librispeech} (960\,h) and for French, we tested a wav2vec model trained on the French Voxpopuli dataset \cite{wang-etal-2021-voxpopuli} (4.5k\,h). The results of these models compared to ours and MFCCs can be seen in Figure \ref{fig:pretrained}. Their different ABX scores can also be seen in Table \ref{tab:ABX_pret}. Models trained on English are evaluated on English-speaking participants (and English contrast for the ABX scores), and same for French.
\label{sec:pretrained}

\begin{table}
\centering
\begin{tabular}{l|ll|ll|l}
Models & \multicolumn{2}{l}{Zerospeech}         & \multicolumn{2}{l}{WorldVowels}      & PA\\
& FR & EN& FR& EN& EN\\
w2v-nat    & \textbf{0.88} & 0.88 & 0.71 & 0.83& 0.84 \\
w2v-pret    & 0.85&0.86 &0.69 &0.84& 0.86 \\
hub-nat     & 0.87 & 0.87& \textbf{0.76} & 0.83 & 0.82   \\
hub-pret     &- &\textbf{0.89} &- &\textbf{0.89} &\textbf{0.90} \\
mfccs  & 0.76& 0.77        & 0.73        & 0.76        & 0.88
\end{tabular}
\caption{ABX scores of our self-supervised models (-nat) compared to pretrained ones (-pret). Best results for each subset is in bold}
\label{tab:ABX_pret}
\end{table}

\section{Testing DeepSpeech using orthographic transcriptions}
\label{sec:DeepSpeech_ortho}
\begin{figure}
   \centering
    \includegraphics[trim ={0.2cm 0cm 0cm 1cm}, clip,scale = 0.51]{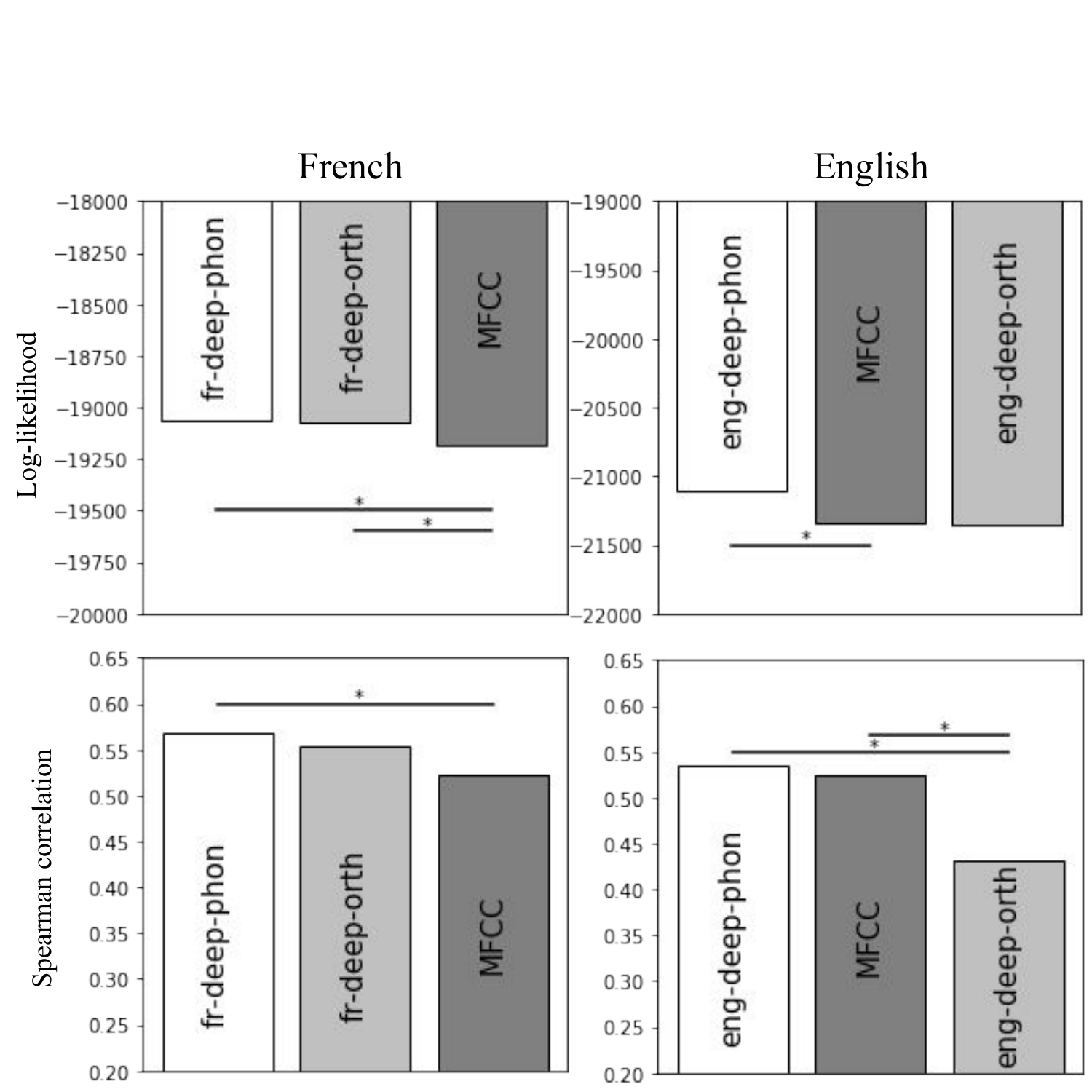}
    \caption{Results of DeepSpeech trained on phonemic transcriptions (phon) or orthographic (orth), compared with MFCCs. Log-likelihood values (top: shorter bars are better) and Spearman correlation (bottom: taller bars are better) for French (\emph{left}) and English participants (\emph{right}). Stars indicate that the pairwise difference is significant.}
    \label{fig:DeepSpeech_comp}
\end{figure}

\begin{figure}
    \centering
    \includegraphics[scale=0.10]{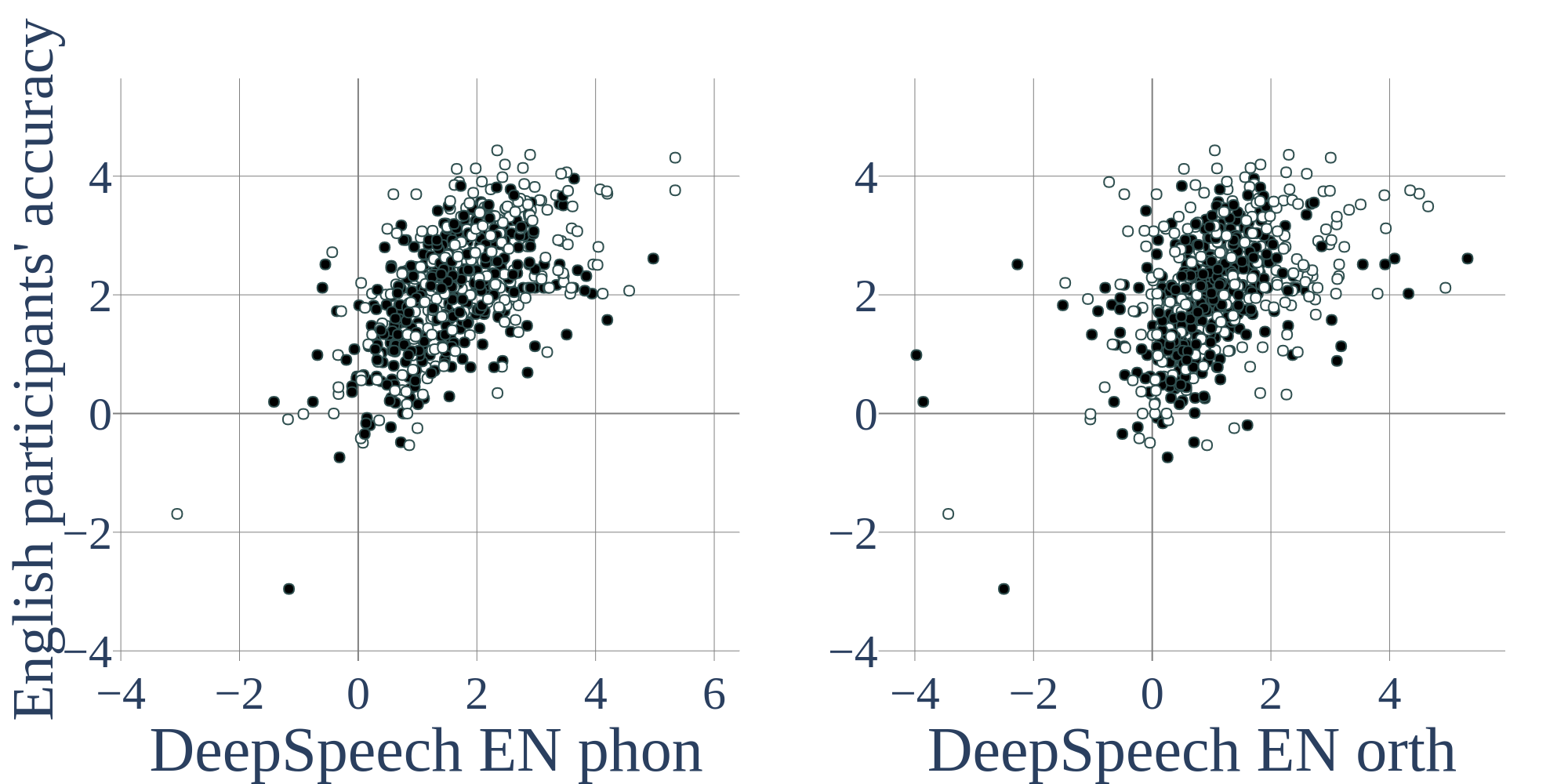}
    \caption{Average of English listeners' results (higher: better discrimination) against average $\delta$ from (\textbf{left}) supervised reference trained on phonemic transcriptions (\textbf{right}) trained on orthographic transcriptions. Each point is a contrast. Measures are normalized by dividing by standard deviation over the entire data set. Black circles are non-native contrasts, white ones are native (English).}
    \label{fig:DeepSpeech_compa_hum}
\end{figure}

We tested two kinds of supervised references: one trained to produce phonemic transcriptions (the one used in the main article) and another trained to produce orthographic transcriptions. In general, training on phonemic transcriptions led the internal representations of the model to be closer to humans' perceptual space, as it can be seen in Figure \ref{fig:DeepSpeech_comp}. A comparison of English-speaking participants' discrimination ability and the two supervised models' $\Delta$-values can also be seen in Figure \ref{fig:DeepSpeech_compa_hum}. Models trained on phonemic transcriptions are better at predicting human behaviour than the ones trained on orthographic transcriptions. These results highlight on the one hand the impact of the labels used during supervised training, which can lead to non human-like speech representational space, and on the other hand the fact that humans probably use informations more similar to phoneme categories than possible orthographic transcriptions during a discrimination task.

The amount of training data may also play a role, as large training sets could lead to ``overfitting,'' in a loose sense, to fine ``superhuman'' acoustic details of phone classification.  Appendix \ref{sec:pretrained} shows that training size does \emph{not} have this effect on the self-supervised models studied here. We leave analysis of the supervised case for future work.

\end{document}